\documentclass[letterpaper,10pt,conference]{ieeeconf}
\IEEEoverridecommandlockouts
\usepackage[T1]{fontenc}
\usepackage{mathptmx}
\usepackage{amsmath,amssymb}
\usepackage{graphicx}
\newsavebox{\marginfigurebox}
\newcommand{\marginsafegraphic}[2][]{%
  \sbox{\marginfigurebox}{\includegraphics[#1]{#2}}%
  \raisebox{0pt}[\ht\marginfigurebox][\dp\marginfigurebox]{%
    \makebox[\wd\marginfigurebox][c]{%
      \resizebox{!}{\dimexpr\ht\marginfigurebox-6bp\relax}{\usebox{\marginfigurebox}}}}%
}
\usepackage{booktabs,array,multirow}
\usepackage[sort,compress]{cite}

\usepackage{xcolor}
\usepackage{url}
\newcommand{\ours}{BAAT}

\newcommand{\hist}{\mathbf A_{t-1}^{\mathrm{hist}}}
\newcommand{\Zq}{\mathbf Z_{q,t}}
\title{\LARGE\bf Behavior-Aligned Action Tokenization for Robot Policy Learning}
\author{Junbo Dong$^{1}$, Ze Chen$^{2}$, Zhendong Xie$^{1}$, Junjie Li$^{1,*}$,\\
Lixin Xu$^{3}$, Xuemin Chi$^{4}$, Yiming Song$^{1}$, and Zhaoyuan Ma$^{1}$%
\thanks{$^{1}$Junbo Dong, Zhendong Xie, Junjie Li, Yiming Song, and Zhaoyuan Ma are with Southern University of Science and Technology.}%
\thanks{$^{2}$Ze Chen is with Dexmal.}%
\thanks{$^{3}$Lixin Xu is with National University of Singapore.}%
\thanks{$^{4}$Xuemin Chi is with Zhejiang University.}%
\thanks{$^{*}$Corresponding author: Junjie Li.}%
}
\begin{document}
\maketitle
\thispagestyle{empty}
\pagestyle{empty}
\raggedbottom

\begin{abstract}
Autoregressive robot policies learn continuous control by predicting discrete action tokens from observations. Different tasks often share local motions, yet behavioral correspondence across demonstrations receives limited explicit supervision in existing tokenizers. Motions with different timing can therefore lack a shared representation despite following similar patterns. We propose Behavior-Aligned Action Tokenization (BAAT), which uses soft dynamic time warping (Soft-DTW) to select corresponding action chunks and aligns their quantized coordinates jointly with reconstruction. This objective encourages similar motions across tasks to occupy nearby quantized representations while retaining executable action detail. A history-conditioned diffusion decoder reconstructs continuous action chunks from these tokens, and a downstream autoregressive policy learns to predict them. We evaluate BAAT on selected tasks from three simulation benchmarks and two real robot tasks. BAAT achieves a mean simulation success rate of approximately 45.2\%, exceeding OAT by approximately 7.2 percentage points. In the controlled LIBERO-All alignment ablation, policy success rises from 70.2\% to 79.0\% while trajectory replay success decreases. These results support behavioral correspondence as supervision for organizing shared motion structure in action tokenizers and improving downstream robot policy learning.
\end{abstract}

\section{Introduction}
\label{sec:intro}
Robot policies that generalize across tasks and environments must learn reusable behavior from diverse experience~\cite{oxe}. Recent vision-language-action (VLA) models demonstrate how pretrained representations and heterogeneous training data can support this generalization~\cite{openvla,pi05}. Robot control also depends on how continuous actions are represented. For autoregressive policies, an action tokenizer converts demonstration action chunks into discrete prediction targets. FAST and OAT show that compression and token ordering can improve the efficiency of this interface~\cite{fast,oat}, motivating attention to target structure.

Recent work also explores learned quantization for policy prediction in VQ-BeT~\cite{vqbet} and VQ-VLA~\cite{vqvla}, token stability and policy optimization in ActionCodec~\cite{actioncodec}, and multimodal alignment in X-Tokenizer~\cite{xtokenizer}. We investigate behavioral correspondence across demonstrations as a complementary, action-only source of tokenizer supervision.

Different tasks often share local motions, such as reaching forward, lifting an object, or transferring it laterally. These patterns recur with different timing. Play-LMP~\cite{playlmp} organizes play behaviors in a latent space for reuse across goals, while QueST~\cite{quest} and PRISE~\cite{prise} learn temporal action abstractions that support downstream learning across tasks. Chunk reconstruction alone does not specify which motions across demonstrations should occupy nearby representations. Motion correspondence can organize chunks across task boundaries. We ask: \emph{can behavioral correspondence supervise a quantized action representation that supports multi-task policy learning?}

Two considerations guide our approach. First, correspondence should accommodate timing differences: pointwise comparison can separate similar motions with different speeds or pauses. Temporal Cycle-Consistency Learning~\cite{tcc} and Learning by Aligning Videos in Time~\cite{lav} use temporal correspondence to supervise video representations; BAAT selects corresponding pairs from action chunks. Second, alignment should act on the quantized representations supplying policy targets while retaining executable detail. We combine temporal matching, alignment after quantization, and action reconstruction. The shared structure consists of neighborhoods of similar local motions across tasks; it does not require shared task semantics or identical discrete codes.

We introduce \ours{} (\emph{Behavior-Aligned Action Tokenization}; Fig.~\ref{fig:overview}). Soft dynamic time warping (Soft-DTW)~\cite{softdtw} selects neighboring action chunks within a mini-batch, and behavioral alignment encourages their finite scalar quantization (FSQ)~\cite{fsq} coordinates to remain close. We train the tokenizer jointly with a history-conditioned diffusion decoder, then freeze it to provide categorical targets for an autoregressive policy. Predicted labels map to quantized coordinates for continuous action decoding.

Cross-task retrieval shows corresponding local motions in BAAT's quantized neighborhoods (Fig.~\ref{fig:behavioral-neighborhood}). In a controlled LIBERO-All ablation, alignment raises policy success from 70.2\% to 79.0\% while replay success decreases from 80.5\% to 71.5\% (Table~\ref{tab:alignment}). These observations connect shared motion structure, action fidelity, and control. Evaluations span three simulation benchmarks and two real robot tasks, with BAAT gaining approximately 8.1 percentage points from joint LIBERO training (Table~\ref{tab:joint}).

Our contributions are:
\begin{itemize}
\item We propose BAAT, an action-only tokenizer that uses Soft-DTW behavioral correspondence to align quantized action representations jointly with reconstruction.
\item We identify cross-task motion neighborhoods in BAAT's quantized space and analyze alignment's benefits and limits for action fidelity and policy performance.
\item Across three simulation benchmarks and two real robot tasks, BAAT achieves the highest mean simulation success and largest joint LIBERO training gain among compared methods.
\end{itemize}

\section{Related Work}
\label{sec:related}
\subsection{Action Tokenization for Robot Policies}
Action tokenizers determine both the length and structure of a policy's discrete prediction targets. Direct coordinate binning assigns a categorical label to each action coordinate, so the sequence length grows with action dimension and prediction horizon~\cite{rt1}. FAST~\cite{fast} exploits temporal structure through the discrete cosine transform and compresses the coefficients with byte-pair encoding, providing a compression-based design for these prediction targets.

Learned tokenizers introduce additional objectives for organizing these targets~\cite{vqvae}. OAT~\cite{oat} combines FSQ, nested dropout, and causal attention to order tokens for reconstruction from progressively longer prefixes. ActionCodec~\cite{actioncodec} studies token stability through temporal overlap, alongside vocabulary redundancy, multimodal mutual information, and token independence. Its temporal contrastive objective uses adjacent action chunks as positive pairs in continuous embedding space. X-Tokenizer~\cite{xtokenizer} combines hierarchical residual quantization with foundation-model alignment and future vision-language feature prediction, supplying multimodal and hierarchical supervision. BAAT selects pairs by motion correspondence among mini-batch chunks and constrains their quantized coordinates. This action-only supervision is designed to organize similar motions across task contexts without requiring temporal adjacency or task labels.

\subsection{Behavior-Aware Action Representation}
Behavioral context and dynamics provide supervision for action representations. Act2Vec~\cite{act2vec} learns action embeddings from demonstration context, while Dynamics-Aware Embeddings~\cite{dyne} jointly represents states and action sequences through forward prediction.

Temporal correspondence provides another source of supervision. Time-Contrastive Networks~\cite{tcn} use synchronized observations from different viewpoints; Temporal Cycle-Consistency Learning~\cite{tcc} learns frame correspondence across unsynchronized videos through cycle consistency. Action-centric cycle consistency~\cite{unified} trains latent action representations by recovering sampled actions from original and generated video frames. HiLAM~\cite{hilam} aggregates low-level latent actions over longer temporal sequences to learn high-level skills. BAAT uses measured action chunks to supply a related form of supervision: Soft-DTW selects pairs for behavioral alignment in quantized coordinates.

\subsection{Generative Action Modeling and Reconstruction}
Continuous generative models provide decoders for robot actions. Diffusion Policy~\cite{dp} generates action trajectories through conditional denoising, and Octo~\cite{octo} integrates a diffusion action head into a generalist policy. The $\pi_{0.5}$ model~\cite{pi05} combines discrete action-token training with a continuous flow-matching action expert.

BAAT uses conditional denoising to reconstruct continuous actions from behavior-aligned quantized representations and executed action history.

\section{Method}
\label{sec:method}
\begin{figure*}[t]
\centering
\marginsafegraphic[width=\textwidth]{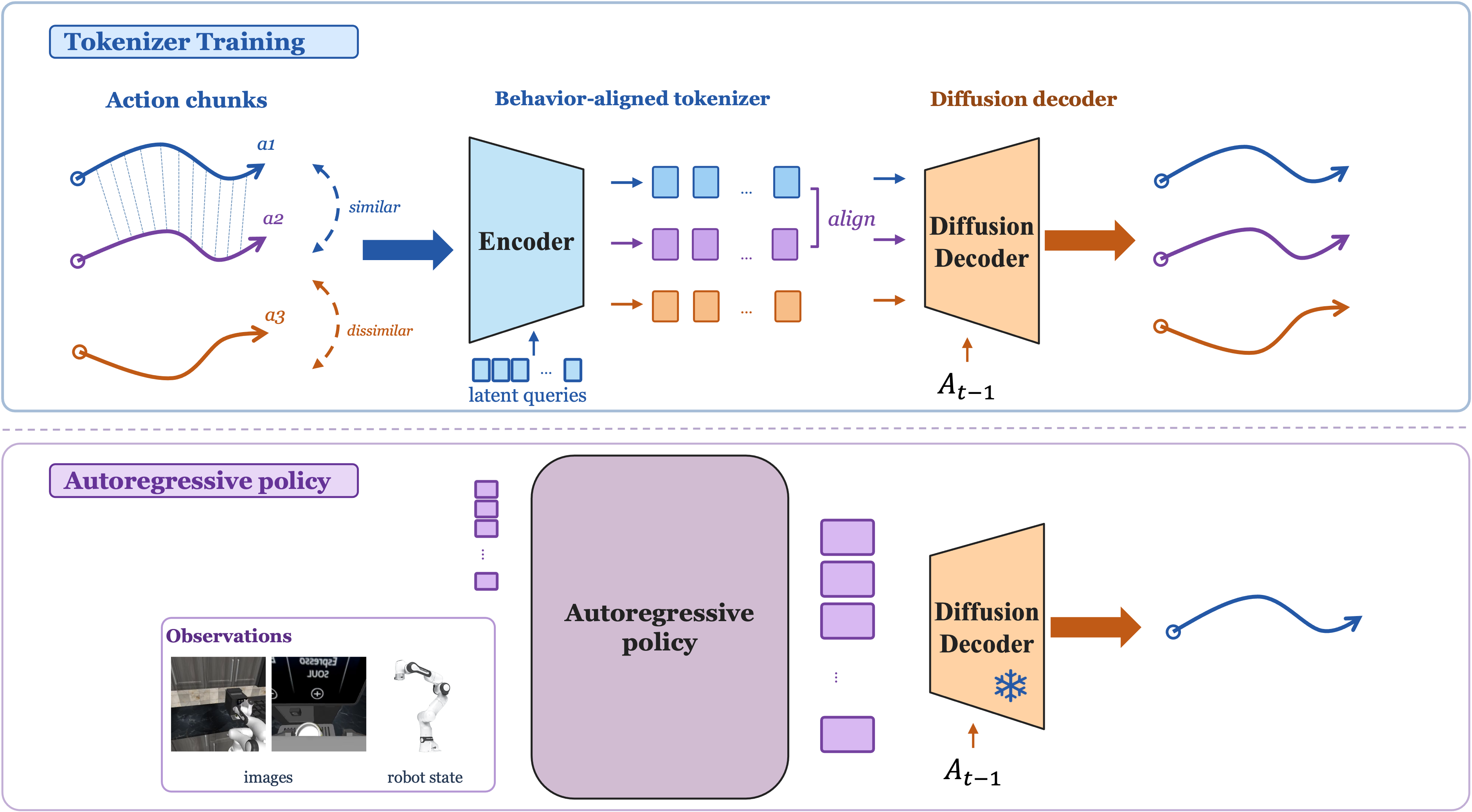}
\caption{BAAT tokenizer training and downstream policy learning. Soft-DTW supplies behavioral correspondence between action chunks; alignment constrains their quantized coordinates jointly with reconstruction. Each latent slot contains four FSQ factors. The diffusion decoder conditions on executed action history, indicated by $A_{t-1}$ in the schematic.}
\label{fig:overview}
\end{figure*}

\subsection{Method Overview}
\label{sec:overview}
BAAT aims to organize similar local motions into shared quantized neighborhoods while retaining the detail needed for execution. It has two training stages (Fig.~\ref{fig:overview}). First, Soft-DTW selects action-chunk pairs as behavioral correspondence supervision. We jointly train an encoder and a history-conditioned diffusion decoder with alignment on the paired quantized coordinates and action reconstruction. Second, we freeze the tokenizer and decoder, tokenize expert chunks offline, and train an autoregressive policy on the resulting categorical labels. At deployment, the decoder reconstructs continuous actions from predicted labels mapped to quantized coordinates and from executed action history.

\subsection{Behavior-Aligned Action Tokenization}
\subsubsection{Action representation}
We use FSQ~\cite{fsq} to represent action chunks in a discrete coordinate space where trajectory neighbors can be aligned. At control time $t$, let $\mathbf A_t=[\mathbf a_t,\ldots,\mathbf a_{t+H-1}]\in\mathbb R^{H\times D_a}$ contain $H$ commands of dimension $D_a$. An encoder $E_\phi$ produces $L$ latent slots of hidden dimension $D$, and a linear projection $P_\omega$ maps each slot to $M$ scalar coordinates. Quantization yields categorical targets $\mathbf C_t$ and their coordinates $\Zq\in\mathbb R^{L\times M}$:
\begin{equation}
\mathbf C_t=Q(P_\omega(E_\phi(\mathbf A_t))),\qquad
\Zq=r(\mathbf C_t).
\label{eq:representation}
\end{equation}
Here $Q$ quantizes each scalar independently, producing a categorical label $c_{t,l}^{(m)}\in\{0,\ldots,K_m-1\}$ for latent slot $l$ and FSQ factor $m$, where $K_m$ is its number of levels. The map $r$ converts each label to its fixed scalar level. The policy predicts $\mathbf C_t$; alignment and decoding use $\Zq$.

\subsubsection{Behavioral alignment}
Similar motions can differ in speed or phase, causing pointwise comparisons to assign large distances to corresponding trajectories. For matching, we represent each action chunk as a normalized motion-increment sequence $\tau_i=[\mathbf u_i^1,\ldots,\mathbf u_i^H]$, where $\mathbf u_i^h$ denotes the motion increment at step $h$ in the platform-specific matching space. We allow temporal warping through the DTW cost~\cite{dtw}:
\begin{equation}
d_{\mathrm{DTW}}(\tau_i,\tau_j)
=\min_{\pi\in\Pi}\sum_{(p,q)\in\pi}
\|\mathbf u_i^p-\mathbf u_j^q\|_2^2,
\label{eq:dtw}
\end{equation}
where $\Pi$ contains monotone, continuous alignment paths. We use GPU-batched Soft-DTW~\cite{softdtw}, the smoothed counterpart of this cost, to select the $K_{\mathrm{pair}}$ lowest-scoring pairs among all $1\leq i<j\leq B$ in a mini-batch of size $B$. The selected set $\mathcal P$ provides weak behavioral correspondence, not task-semantic labels.

Behavioral alignment acts on these pairs after quantization:
\begin{equation}
\mathcal L_{\mathrm{align}}
=\frac{1}{|\mathcal P|LM}
\sum_{(i,j)\in\mathcal P}
\|\mathbf Z_{q,i}-\mathbf Z_{q,j}\|_F^2.
\label{eq:align}
\end{equation}
This loss encourages nearby quantized representations without requiring identical codes. It measures distances between scalar coordinates, not between categorical label identities. Soft-DTW determines which chunks to align; this loss constrains the representation supplied to the policy, while joint reconstruction penalizes the loss of executable detail.

\subsection{History-Conditioned Diffusion Reconstruction}
\label{sec:decoder}
A discrete code can correspond to multiple continuous action realizations, including variations in timing and amplitude. We use a diffusion decoder to model $p_\theta(\mathbf A_t\mid\Zq,\hist)$, where $\hist$ contains actions executed before the current chunk. This history supplies the decoder with execution context from the preceding chunk.

At diffusion step $s$, we corrupt the action chunk as $\widetilde{\mathbf A}_{t,s}=\alpha_s\mathbf A_t+\sigma_s\boldsymbol\epsilon$, where $\boldsymbol\epsilon\sim\mathcal N(\mathbf0,\mathbf I)$ and $(\alpha_s,\sigma_s)$ defines the noise schedule~\cite{ddpm}. The decoder directly predicts the clean chunk $\widehat{\mathbf A}_t=D_\theta(\widetilde{\mathbf A}_{t,s},s,\Zq,\hist)$ and minimizes the masked Smooth-$L_1$ loss:
\begin{equation}
\mathcal L_{\mathrm{rec}}
=\mathbb E_{\mathbf A_t,s,\boldsymbol\epsilon}
\left[\frac{\sum_{h,d}M_{h,d}\,
\rho(\widehat A_{t,h,d}-A_{t,h,d})}
{\sum_{h,d}M_{h,d}}\right],
\label{eq:rec}
\end{equation}
where $M_{h,d}$ identifies valid action elements and $\rho$ is the Smooth-$L_1$ function. With alignment weight $\lambda_{\mathrm{align}}$, the joint tokenizer objective is
\begin{equation}
\mathcal L_{\mathrm{tokenizer}}
=\mathcal L_{\mathrm{rec}}+\lambda_{\mathrm{align}}\mathcal L_{\mathrm{align}}.
\label{eq:total}
\end{equation}
A straight-through gradient approximation~\cite{ste} allows both losses to update the encoder and projection through quantization.

For the 12D bimanual real-robot configuration evaluated in Sec.~\ref{sec:history-continuity}, history is the immediately preceding action $\mathbf a_{t-1}$. We normalize it with the action statistics, encode it with an MLP, and inject it into the decoder's FiLM modulation~\cite{film}. The decoder uses a residual output parameterization: $\widehat{\bar{\mathbf a}}_{t+h}=\widehat{\mathbf r}_{t,h}+\mathbf m\odot\bar{\mathbf a}_{t-1}$, where bars denote normalized actions and the binary mask $\mathbf m$ selects the ten arm joints. Each arm prediction is an offset from the same preceding-action anchor, broadcast across the chunk; adding this anchor recovers absolute targets in normalized space. Inverse normalization then restores raw action units. The two grippers are predicted as absolute targets, and the encoder still receives absolute action chunks. This configuration also uses boundary and overlap losses and a longer prediction horizon.

\subsection{Factorized Autoregressive Policy Learning}
\label{sec:policy}
The frozen tokenizer supplies offline targets $\mathbf C_t$ for expert action chunks. Conditioned on observations $\mathbf o_t$ and an optional task input $\mathbf g_t$, the policy autoregresses over latent slots and predicts their FSQ factors independently in parallel using a $K_m$-class head for each factor $m$. We train with teacher forcing and the factorized negative log-likelihood:
\begin{equation}
\mathcal L_{\mathrm{AR}}=-\frac{1}{LM}
\sum_{l=1}^{L}\sum_{m=1}^{M}
\log p_\psi(c_{t,l}^{(m)*}\mid\mathbf C_t^{<l},\mathbf o_t,\mathbf g_t),
\label{eq:ar}
\end{equation}
where $\psi$ denotes the policy parameters, starred indices are tokenizer-produced targets, and $\mathbf C_t^{<l}$ contains preceding target latent slots during training.

At inference, the policy conditions on its previously generated latent slots. The map $r$ converts the sampled FSQ factor labels to quantized coordinates, which condition the frozen diffusion decoder together with executed action history to generate the next chunk by iterative denoising.

\section{Experiments}
\label{sec:experiments}
We first inspect cross-task behavioral neighborhoods, then evaluate control and action fidelity across training regimes. Simulation ablations test behavioral correspondence supervision and its balance with reconstruction. A separate offline decoder-history comparison provides a supplementary diagnosis of chunk continuity on recorded real-robot data.

\subsection{Experimental Setup}
\subsubsection{Tasks and training regimes}
We use 40 LIBERO~\cite{libero} tasks from LIBERO-10, Spatial, Goal, and Object (10 tasks each), training either one policy per suite (\emph{per-suite training}) or one policy jointly on demonstrations from all four suites (\emph{LIBERO-All}).

\begin{figure}[t]
\centering
\marginsafegraphic[width=\columnwidth]{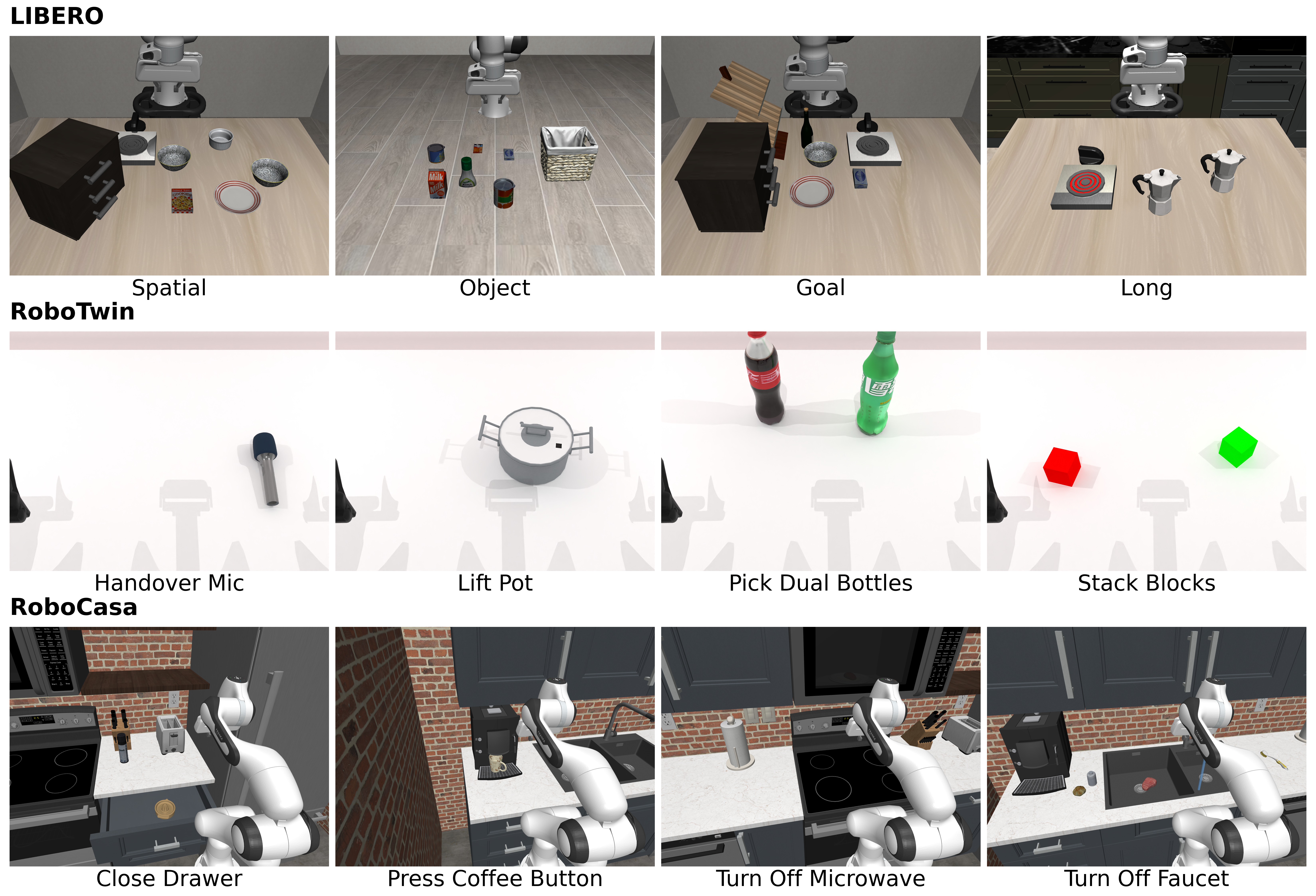}
\caption{Simulation task overview. Top: example scenes from the LIBERO Spatial, Object, Goal, and long-horizon suites. Middle: RoboTwin 2.0 tasks, comprising Handover Mic, Lift Pot, Pick Dual Bottles, and Stack Blocks. Bottom: RoboCasa tasks, comprising Close Drawer, Press Coffee Button, Turn Off Microwave, and Turn Off Faucet.}
\label{fig:benchmarks}
\end{figure}

We also evaluate four RoboCasa~\cite{robocasa} tasks involving kitchen manipulation and four RoboTwin 2.0~\cite{robotwin} tasks involving bimanual control. Each benchmark uses joint training across its selected tasks. Fig.~\ref{fig:benchmarks} illustrates the simulation tasks. Real robot evaluation covers toothbrush organization and object sorting on two bimanual systems. Object sorting requires coordinated control of two 7-DoF arms with two 21-DoF dexterous hands, yielding a 56-DoF action space.

\subsubsection{Baselines and policy training}
Baselines include discrete action interfaces FAST, direct coordinate binning (Bin), and OAT, and continuous-action Diffusion Policy (DP). Discrete interfaces share the downstream autoregressive architecture and training procedure where applicable, retaining tokenizer-specific sequence structures.

BAAT uses $L=16$ latent slots, hidden dimension $D=512$, and $M=4$ FSQ factors per slot with levels $(8,5,5,5)$. Main policy experiments predict $H=20$ action steps and execute the first 16 before observing and replanning, with alignment weight $\lambda_{\mathrm{align}}=0.1$ unless specified otherwise.


\subsubsection{Metrics and Evaluation Protocol}
We report closed-loop task success rate (policy SR). Simulation policy comparisons and ablations are averaged over three random seeds. Each run's final checkpoint is evaluated with 50 rollouts per task; task SRs are averaged equally within each run, then across seeds. Suite and overall benchmark averages equally weight the four suites and three benchmarks, respectively.

Simulation ablations use LIBERO-All joint training; the alignment experiment varies only the alignment weight, keeping all other training and evaluation settings identical.

Simulation tokenizer metrics are reconstruction mean squared error (MSE) and replay SR from executing reconstructed demonstration actions in the original environment. Real robot results report SR and successful trials out of 20 attempts per task and method.

The separate offline diagnostic measures arm reconstruction mean absolute error (MAE) and boundary-jump quantiles in raw joint-target units on recorded real-robot data, without robot execution (Sec.~\ref{sec:history-continuity}).

\begin{figure*}[t]
\centering
\includegraphics[width=0.85\textwidth]{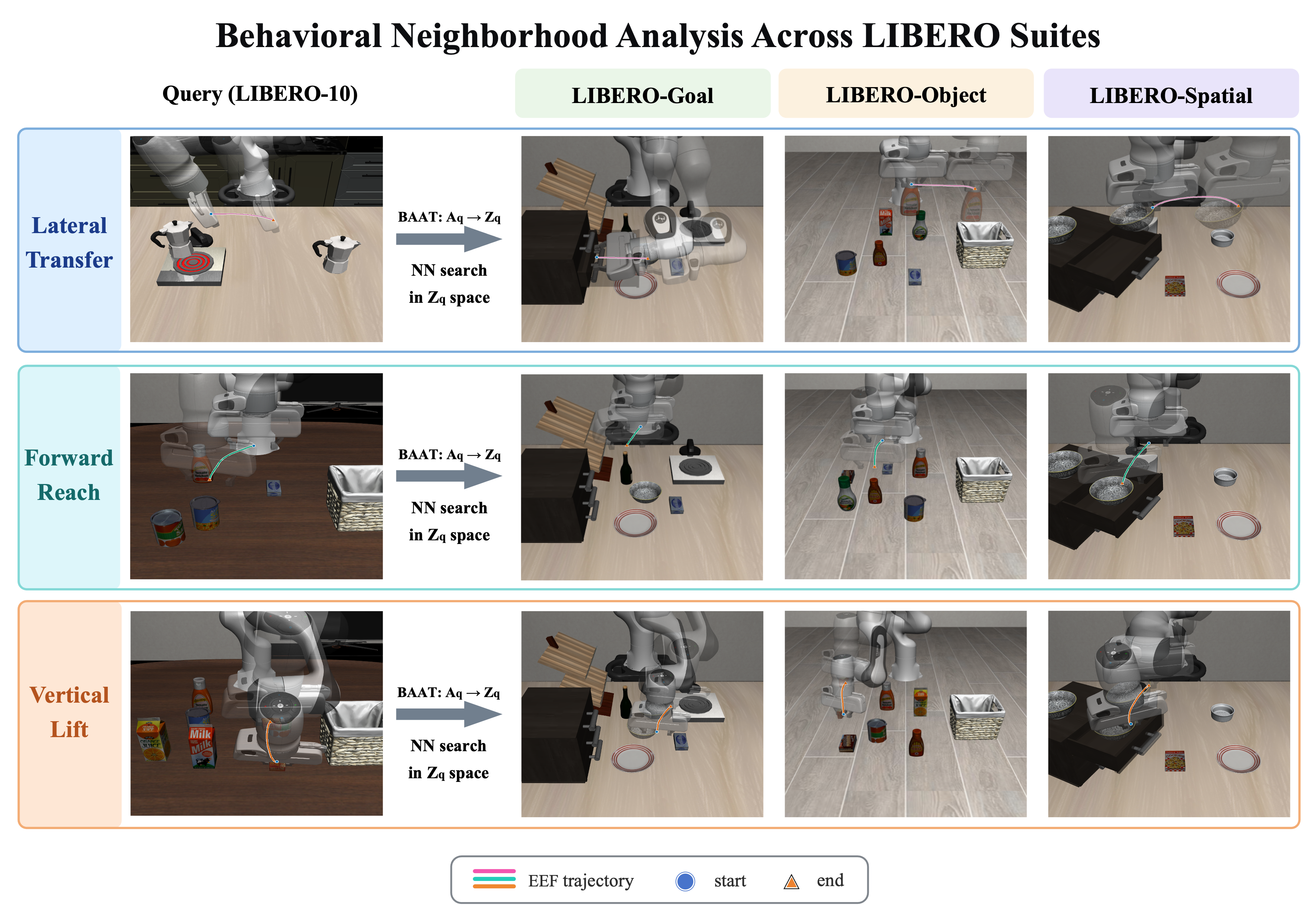}
\caption{Cross-task behavioral neighborhoods. Left: LIBERO-10 queries; right: nearest neighbors from LIBERO-Goal, Object, and Spatial. Rows show lateral transfer, forward reach, and vertical lift; circles and triangles mark trajectory starts and ends. Plots show the retrieved chunks' trajectories, without decoder generation or selection from trajectory plots.}
\label{fig:behavioral-neighborhood}
\end{figure*}

\subsection{Cross-Task Behavioral Neighborhoods}
\label{sec:behavioral-neighborhood}
We ask whether proximity in BAAT's quantized space reflects local motions shared across tasks. For each LIBERO-10 query $\mathbf Z_q$, we retrieve the nearest neighbor independently from LIBERO-Goal, LIBERO-Object, and LIBERO-Spatial using quantized coordinate distance, then inspect the associated action trajectories.

Fig.~\ref{fig:behavioral-neighborhood} shows three recurring motion patterns amid different objects and scene layouts. The lateral-transfer paths share a predominantly sideways displacement, with both nearly straight and curved paths. The forward-reach paths extend toward different parts of the workspace, varying in curvature and endpoints. The vertical-lift paths retain an upward trend while differing in their lateral components. Thus, nearby representations associate corresponding dominant motions while retaining trajectory differences. This local correspondence does not require tasks to share the complete manipulation goal.

Retrieval supplies qualitative evidence for shared local motion structure; the joint-training comparison and alignment ablations examine complementary control benefits and fidelity trade-offs (Secs.~\ref{sec:joint} and~\ref{sec:ablations}). The current retrieval does not isolate alignment's role in forming these neighborhoods or establish independently callable skills.

\subsection{Downstream Policy Performance}
\subsubsection{Simulation benchmarks}
\begin{table}[h!]
\centering
\caption{Success rates (\%) under joint training.}
\label{tab:simulation}
\setlength{\tabcolsep}{3pt}
\renewcommand{\arraystretch}{1.12}
\begin{tabular*}{\columnwidth}{@{\extracolsep{\fill}}lrrrr@{}}
\toprule
Method & LIBERO-All & RoboCasa & RoboTwin~2.0 & Avg.\\
\midrule
FAST & 63.5 & 4.3 & 5.3 & 24.4\\
OAT & 76.5 & 23.3 & 14.2 & 38.0\\
Bin & 45.2 & 18.7 & 11.8 & 25.2\\
DP & 62.9 & 13.8 & 25.7 & 34.1\\
\ours{} (ours) & \textbf{79.0} & \textbf{29.2} & \textbf{27.3} & \textbf{45.2}\\
\bottomrule
\end{tabular*}
\end{table}
BAAT achieves the highest mean policy SR among the compared methods on each simulation benchmark (Table~\ref{tab:simulation}), with an unweighted mean of approximately 45.2\%, exceeding OAT by approximately 7.2 percentage points. On LIBERO-All, it reaches 79.0\%, approximately 2.5 points above OAT.
These results support the use of behavior-aligned action tokens as a policy interface across the evaluated manipulation settings.

\subsubsection{Per-suite versus joint training}
\label{sec:joint}
BAAT benefits most from learning across suites (Table~\ref{tab:joint}): joint training raises its mean policy SR from 70.9\% to 79.0\%, yielding the highest joint score and the largest gain, approximately 8.1 percentage points. FAST, OAT, and Bin gain 1.6, 1.3, and 2.5 points, respectively. BAAT's per-suite average is below OAT and DP, and it trails FAST, OAT, and DP on independently trained LIBERO-10. Its advantage therefore emerges most clearly when demonstrations from different suites are learned together.

\begin{table}[h!]
\centering
\caption{LIBERO success rates (\%): per-suite versus joint training.}
\label{tab:joint}
\setlength{\tabcolsep}{2pt}
\renewcommand{\arraystretch}{1.12}
\begin{tabular*}{\columnwidth}{@{\extracolsep{\fill}}lrrrrrr@{}}
\toprule
Method & LIBERO-10 & Spatial & Goal & Object & Suite avg. & Joint\\
\midrule
FAST & 32.5 & 67.8 & 75.8 & 71.5 & 61.9 & 63.5\\
OAT & 43.7 & 78.3 & 85.3 & \textbf{93.6} & \textbf{75.2} & 76.5\\
Bin & 15.2 & 39.7 & 47.6 & 68.1 & 42.7 & 45.2\\
DP & \textbf{66.0} & 73.5 & 79.3 & 75.8 & 73.7 & 62.9\\
\ours{} (ours) & 26.0 & \textbf{81.9} & \textbf{85.6} & 90.0 & 70.9 & \textbf{79.0}\\
\bottomrule
\end{tabular*}
\end{table}

BAAT's alignment objective encourages corresponding motions from different tasks to occupy nearby quantized coordinates, providing shared structure in the policy's action targets. The cross-task motion neighborhoods in Fig.~\ref{fig:behavioral-neighborhood} illustrate this structure. Joint training also changes demonstration volume and diversity, so this comparison does not isolate the contribution of alignment. Taken together, the neighborhoods and the larger joint-training gain support the design motivation of organizing shared local motions for multi-task policy learning.

\begin{figure*}[t]
\centering
\begin{minipage}[b]{\columnwidth}
\centering\marginsafegraphic[width=\linewidth,height=0.56\linewidth]{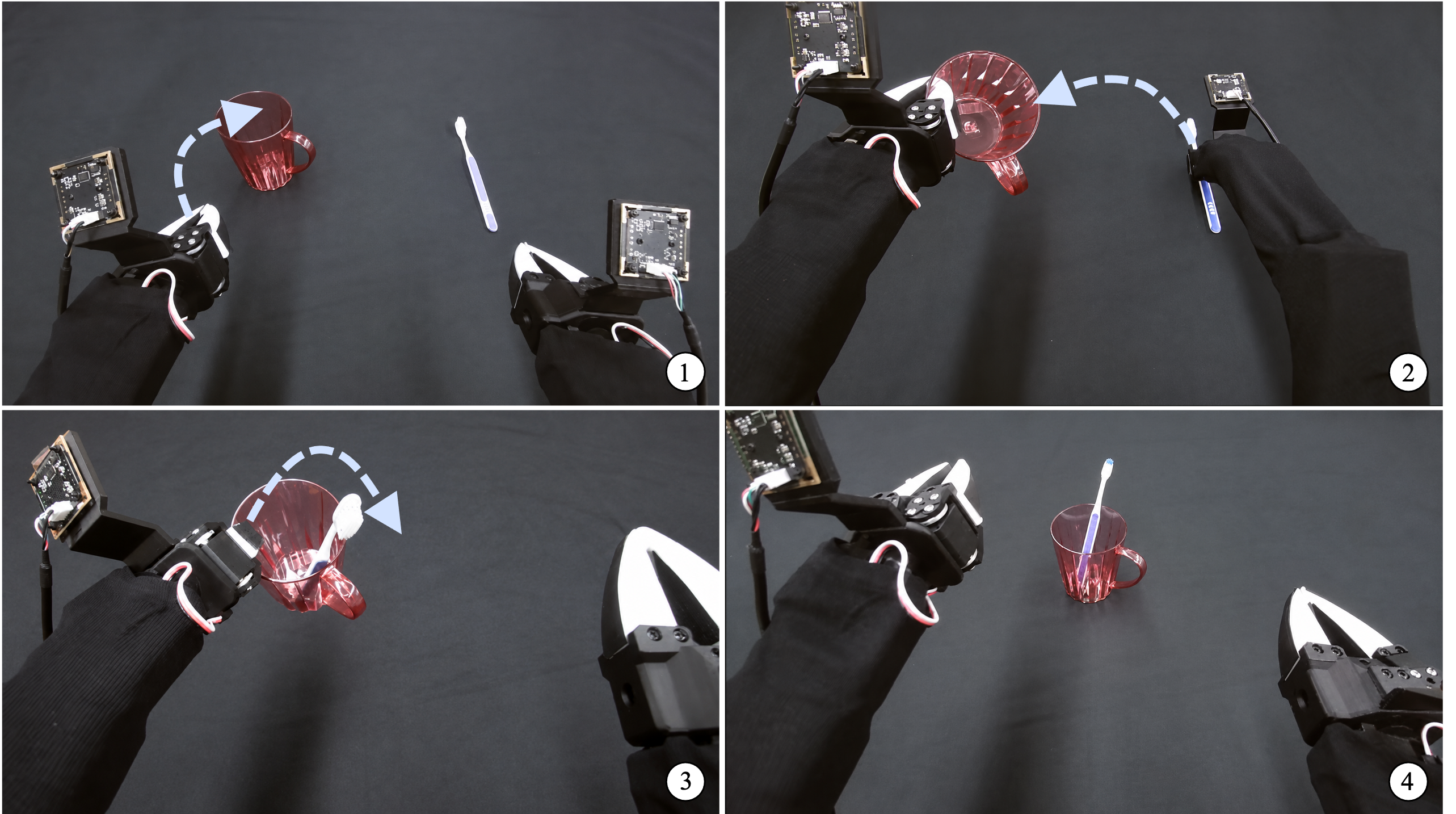}\\[-1pt]
{\footnotesize (a) Toothbrush task}
\end{minipage}\hfill
\begin{minipage}[b]{\columnwidth}
\centering\marginsafegraphic[width=\linewidth,height=0.56\linewidth]{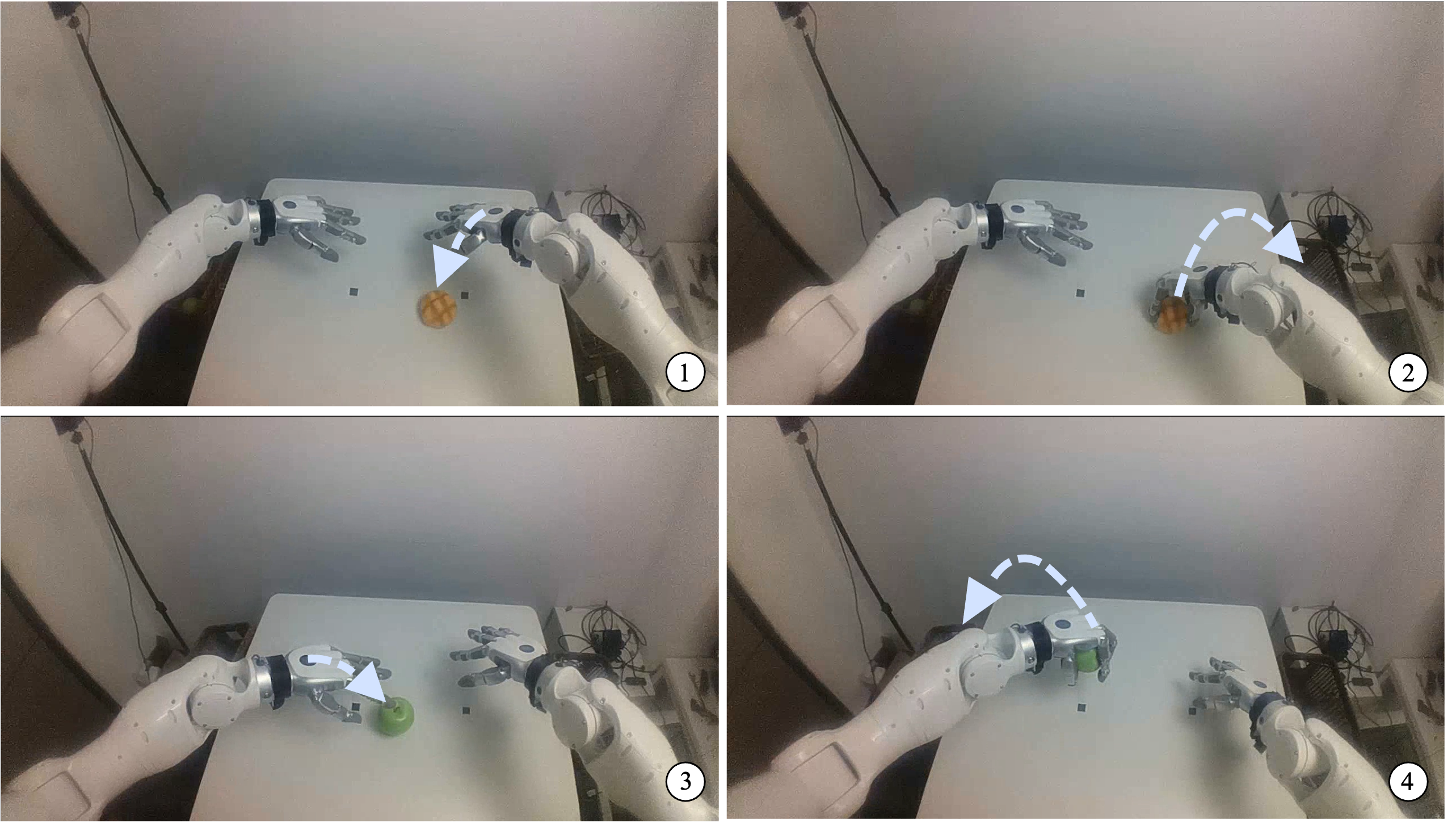}\\[-1pt]
{\footnotesize (b) Object sorting}
\end{minipage}
\caption{Real robot evaluation setups. The two tasks use different bimanual systems; object sorting includes dexterous-hand control.}
\label{fig:robots}
\end{figure*}

\subsubsection{Action Reconstruction}
\begin{table}[h!]
\centering
\caption{Action reconstruction on LIBERO-All.}
\label{tab:reconstruction}
\begin{tabular}{lrr}
\toprule
Tokenizer & Recon. MSE & Replay SR (\%)\\
\midrule
FAST & $5.408\times10^{-5}$ & 90.25\\
Bin & $4.170\times10^{-6}$ & 93.0\\
OAT & $2.274\times10^{-3}$ & 58.5\\
\ours{} (ours) & $5.773\times10^{-4}$ & 71.5\\
\bottomrule
\end{tabular}
\end{table}
\label{sec:reconstruction}
Table~\ref{tab:reconstruction} reports action reconstruction MSE and replay SR on LIBERO-All. Reconstruction starts from tokens encoded from demonstration actions, whereas closed-loop control requires the policy to predict tokens from observations and act on the resulting states. Reconstruction metrics therefore evaluate only part of the deployed interface and are insufficient for choosing a policy interface. These metrics do not directly measure whether BAAT's tokens are easier to predict.

\subsubsection{Real robot manipulation}
\label{sec:real}
Across 20 trials per method and task, BAAT achieves the highest observed success rate on toothbrush organization (14/20) and ties DP on object sorting (9/20; Table~\ref{tab:real}, setups in Fig.~\ref{fig:robots}). These results show that the learned token interface supports closed-loop execution on both real bimanual systems. All methods have lower success rates on the 56-DoF sorting task, which also differs in hardware and manipulation demands, preventing attribution to action dimension alone.

\begin{table}[h!]
\centering
\caption{Real robot success: percentage (successful trials/20).}
\label{tab:real}
\begin{tabular}{lrr}
\toprule
Policy interface & Toothbrush & Object sorting\\
\midrule
FAST & 35 (7/20) & 20 (4/20)\\
OAT & 65 (13/20) & 20 (4/20)\\
Bin & 20 (4/20) & 15 (3/20)\\
DP & 55 (11/20) & 45 (9/20)\\
\ours{} (ours) & 70 (14/20) & 45 (9/20)\\
\bottomrule
\end{tabular}
\end{table}

Real robot deployment showed abrupt chunk transitions despite smooth within-chunk motion; pronounced transition jitter was absent in our simulation evaluations. This motivates the offline comparison in Sec.~\ref{sec:history-continuity}.

\subsubsection{Policy optimization}
\begin{figure}[h!]
\centering
\includegraphics[width=\columnwidth]{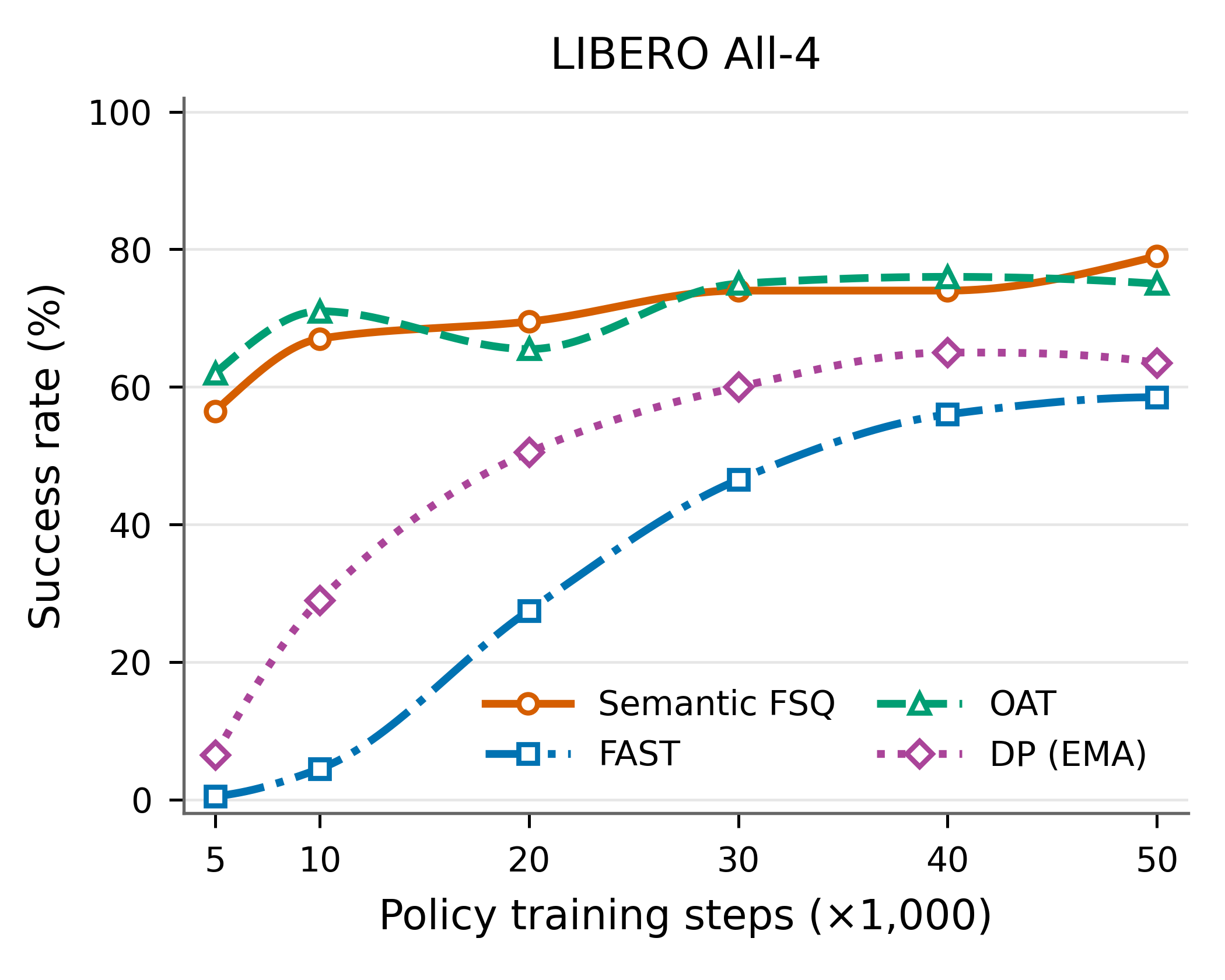}
\caption{LIBERO-All policy SR during one training run per method. The endpoints represent individual runs and need not match the three-seed means in Table~\ref{tab:simulation}. ``LIBERO All-4'' denotes the four-suite setting; ``Semantic FSQ'' denotes BAAT and ``DP (EMA)'' denotes DP.}
\label{fig:convergence}
\end{figure}

In the individual runs in Fig.~\ref{fig:convergence}, BAAT reaches higher policy SR than FAST and DP at early checkpoints, while OAT remains stronger at several early and intermediate checkpoints. BAAT's gains thus depend on the training stage. These single-run curves do not establish generally faster convergence; they complement the final-checkpoint comparison averaged over three seeds in Table~\ref{tab:simulation}, which supports BAAT's benefit in attained policy performance.

\subsection{Reconstruction With and Without Decoder History}
\label{sec:history-continuity}
We compare tokenizer configurations with and without decoder history on recorded bimanual demonstrations: 100 episodes at 30\,Hz with 12D absolute joint targets (five joints and one gripper per arm). Both use 20,000 training steps, batch size 512, and 16 FSQ latent slots. Offline reconstruction uses ten training trajectories and tokens from each configuration's own encoder and quantizer applied to ground-truth actions. History is the ground-truth action immediately preceding each window, without feedback from previous predictions or robot execution.

Starting at frame 16, we sample complete windows every $S=16$ steps within each episode and decode with 27 deterministic diffusion steps. For consecutive reconstructed chunks in raw action units, we define boundary jump as
\begin{equation}
J_k=\left\|\widehat{\mathbf A}_{k+1}[0,\mathcal I]
-\widehat{\mathbf A}_{k}[S-1,\mathcal I]\right\|_2,
\label{eq:boundary-jump}
\end{equation}
where $\mathcal I$ selects the ten arm joints, excluding grippers. Quantiles pool valid boundaries and measure joint-target changes without division by the frame interval. Configurations without/with history yield 577/571 complete windows and 567/561 boundaries. Arm MAE averages elementwise absolute reconstruction errors over each configuration's full prediction windows.

\begin{table}[h!]
\vspace*{6bp}
\centering
\caption{Offline reconstruction with and without decoder history on real-robot data (raw joint-target units; lower is better).}
\label{tab:history-continuity}
\setlength{\tabcolsep}{3pt}
\renewcommand{\arraystretch}{1.12}
\begin{tabular*}{\columnwidth}{@{\extracolsep{\fill}}lrr@{}}
\toprule
Metric & Without history & With history\\
\midrule
Boundary jump $p_{50}$ & 2.505 & 1.367\\
Boundary jump $p_{95}$ & 6.359 & 3.203\\
Boundary jump $p_{99}$ & 8.767 & 4.496\\
Arm reconstruction MAE & 0.4941 & 0.2774\\
\bottomrule
\end{tabular*}
\vspace{-6bp}
\end{table}

The configuration with history reduces boundary-jump $p_{95}$ by approximately 49.6\% and arm MAE by 43.9\% (Table~\ref{tab:history-continuity}), bringing boundary $p_{95}$ closer to the ground-truth reference (2.926/2.940 for the two window sets). Within-chunk one-step jump $p_{95}$ changes much less, from 2.760 to 2.679. The larger change at boundaries indicates that the continuity improvement is concentrated at the joins between reconstructed chunks, where the preceding-action anchor supplies context. The lower arm MAE also shows improved target fidelity, rather than a reduction in jumps alone. Gripper MAE does not improve (0.1477 to 0.1504), limiting the observed reconstruction benefit to the arm channels.

These results diagnose offline reconstruction on training trajectories with ground-truth history. Auxiliary losses, prediction horizons, and window coverage also differ (Sec.~\ref{sec:decoder}), so the gains characterize the combined configuration and cannot be assigned entirely to history input. They do not measure continuity under predicted history or closed-loop performance.

\subsection{Ablation Studies}
\label{sec:ablations}
\begingroup
\makeatletter
\def\@IEEEsectpunct{\ \,}
\makeatother
\subsubsection{How does alignment strength affect policy performance?}
\label{sec:strength}
Varying only $\lambda_{\mathrm{align}}$ tests the balance between behavioral correspondence supervision and reconstruction (Table~\ref{tab:alignment}). Increasing the weight from zero to 0.1 raises policy SR by 8.8 percentage points to 79.0\% while reducing replay SR by 9.0 points. This controlled change directly supports a policy benefit from moderate alignment despite reduced replay fidelity.

\begin{table}[h!]
\centering
\caption{Alignment-weight ablation on LIBERO-All (SR in \%).}
\label{tab:alignment}
\setlength{\tabcolsep}{3pt}
\begin{tabular}{lrrr}
\toprule
$\lambda_{\mathrm{align}}$ & Recon. MSE & Replay SR & Policy SR\\
\midrule
0 & $4.672\times10^{-4}$ & \textbf{80.5} & 70.2\\
0.05 & $5.101\times10^{-4}$ & 76.3 & 73.0\\
0.1 (ours) & $5.773\times10^{-4}$ & 71.5 & \textbf{79.0}\\
0.3 & $9.186\times10^{-4}$ & 65.5 & 57.9\\
0.5 & $1.157\times10^{-2}$ & 0.0 & 0.0\\
\bottomrule
\end{tabular}
\end{table}

Both metrics decline at 0.3 and fall to zero at 0.5. This deterioration is consistent with stronger alignment conflicting with action detail needed for execution, but failure rates alone do not diagnose representation collapse. Among the tested weights, 0.1 achieves the highest policy SR.

\subsubsection{Does temporal matching improve behavioral alignment?}
\label{sec:pairablation}
We compare pointwise $L_2$ matching with Soft-DTW, using no alignment as a reference (Table~\ref{tab:pairs}). Soft-DTW improves policy and replay SR over pointwise matching by 6.0 and 6.9 percentage points, respectively. These gains are consistent with temporal matching accommodating timing offsets between similar motions. This comparison does not directly measure pair quality, and replay SR remains below the no-alignment reference.

\begin{table}[h!]
\centering
\caption{Trajectory matching on LIBERO-All (SR in \%).}
\label{tab:pairs}
\setlength{\tabcolsep}{3pt}
\begin{tabular}{lrrr}
\toprule
Pair construction & Recon. MSE & Replay SR & Policy SR\\
\midrule
No alignment & $4.672\times10^{-4}$ & \textbf{80.5} & 70.2\\
Pointwise $L_2$ & $1.051\times10^{-3}$ & 64.6 & 73.0\\
Soft-DTW (ours) & $5.773\times10^{-4}$ & 71.5 & \textbf{79.0}\\
\bottomrule
\end{tabular}
\end{table}

\subsubsection{How does decoder choice affect downstream control?}
\label{sec:decoderablation}
We compare a deterministic Transformer~\cite{transformer}, diffusion, and flow matching~\cite{flowmatching} to assess decoder choice across reconstruction and control (Table~\ref{tab:decoders}). Each metric favors a different decoder: Transformer minimizes reconstruction MSE, flow matching maximizes replay SR, and diffusion achieves the highest policy SR at 79.0\%, leading by 9.6 and 4.3 percentage points over Transformer and flow matching, respectively.

\begin{table}[h!]
\centering
\caption{Decoder comparison on LIBERO-All (SR in \%).}
\label{tab:decoders}
\setlength{\tabcolsep}{3pt}
\begin{tabular}{lrrr}
\toprule
Decoder & Recon. MSE & Replay SR & Policy SR\\
\midrule
Transformer & $\mathbf{2.413\times10^{-4}}$ & 80.6 & 69.4\\
Diffusion (ours) & $5.773\times10^{-4}$ & 71.5 & \textbf{79.0}\\
Flow matching & $2.444\times10^{-4}$ & \textbf{82.4} & 74.7\\
\bottomrule
\end{tabular}
\end{table}

Because the encoder, quantizer, and decoder are trained jointly, decoder choice affects both the learned representation and its reconstruction. These results compare complete tokenizer configurations, rather than decoder replacements in a fixed quantized space, and do not isolate a particular diffusion property as the source of the policy gain.

\par
\endgroup

\begin{samepage}
\section{Conclusion}
\label{sec:conclusion}
We presented BAAT, an action-only tokenizer that uses behavioral correspondence to organize shared motion structure for multi-task policy learning. BAAT aligns quantized action representations using Soft-DTW correspondence jointly with reconstruction through a history-conditioned diffusion decoder. Cross-task retrieval reveals neighborhoods of similar local motions, while controlled ablations show that moderate alignment improves policy success despite reduced replay success; excessive alignment degrades both. Evaluations on three simulation benchmarks and two real robot tasks demonstrate BAAT's effectiveness, with the highest mean simulation success and largest joint LIBERO training gain among the compared methods. These findings support behavioral correspondence as a useful source of supervision for action tokenization.

An important direction for future work is to use BAAT tokens as prediction targets for pretraining the VLM backbone of a VLA model. Inspired by the discrete action supervision in $\pi_{0.5}$~\cite{pi05} and its knowledge insulation (KI) extension~\cite{ki}, we aim to train the backbone with behavior-aligned tokens while blocking gradients from a continuous action expert. This would allow us to test whether shared motion structure in the token targets improves representation learning and generalization in VLA models.
\end{samepage}


\bibliographystyle{IEEEtran}
\bibliography{references}

\begin{thebibliography}{10}
\providecommand{\url}[1]{#1}
\csname url@rmstyle\endcsname
\providecommand{\newblock}{\relax}
\providecommand{\bibinfo}[2]{#2}
\providecommand\BIBentrySTDinterwordspacing{\spaceskip=0pt\relax}
\providecommand\BIBentryALTinterwordstretchfactor{4}
\providecommand\BIBentryALTinterwordspacing{\spaceskip=\fontdimen2\font plus
\BIBentryALTinterwordstretchfactor\fontdimen3\font minus
  \fontdimen4\font\relax}
\providecommand\BIBforeignlanguage[2]{{%
\expandafter\ifx\csname l@#1\endcsname\relax
\typeout{** WARNING: IEEEtran.bst: No hyphenation pattern has been}%
\typeout{** loaded for the language `#1'. Using the pattern for}%
\typeout{** the default language instead.}%
\else
\language=\csname l@#1\endcsname
\fi
#2}}

\bibitem{oxe}
{Open X-Embodiment Collaboration}, ``{Open X-Embodiment}: Robotic learning
  datasets and {RT-X} models,'' \emph{arXiv preprint arXiv:2310.08864}, 2023.

\bibitem{openvla}
M.~J. Kim, K.~Pertsch, S.~Karamcheti, T.~Xiao, A.~Balakrishna, S.~Nair,
  R.~Rafailov, E.~Foster, G.~Lam, P.~Sanketi, Q.~Vuong, T.~Kollar,
  B.~Burchfiel, R.~Tedrake, D.~Sadigh, S.~Levine, P.~Liang, and C.~Finn,
  ``{OpenVLA}: An open-source vision-language-action model,'' \emph{arXiv
  preprint arXiv:2406.09246}, 2024.

\bibitem{pi05}
{Physical Intelligence}, K.~Black, N.~Brown, J.~Darpinian, K.~Dhabalia,
  D.~Driess, A.~Esmail, M.~Equi, C.~Finn, N.~Fusai, M.~Y. Galliker, D.~Ghosh,
  L.~Groom, K.~Hausman, B.~Ichter, S.~Jakubczak, T.~Jones, L.~Ke, D.~LeBlanc,
  S.~Levine, A.~Li-Bell, M.~Mothukuri, S.~Nair, K.~Pertsch, A.~Z. Ren, L.~X.
  Shi, L.~Smith, J.~T. Springenberg, K.~Stachowicz, J.~Tanner, Q.~Vuong,
  H.~Walke, A.~Walling, H.~Wang, L.~Yu, and U.~Zhilinsky, ``{$\pi_{0.5}$}: A
  vision-language-action model with open-world generalization,'' \emph{arXiv
  preprint arXiv:2504.16054}, 2025.

\bibitem{fast}
K.~Pertsch, K.~Stachowicz, B.~Ichter, D.~Driess, S.~Nair, Q.~Vuong, O.~Mees,
  C.~Finn, and S.~Levine, ``{FAST}: Efficient action tokenization for
  vision-language-action models,'' \emph{arXiv preprint arXiv:2501.09747},
  2025.

\bibitem{oat}
C.~Liu, X.~Han, J.~Gao, Y.~Zhao, H.~Chen, and Y.~Du, ``{OAT}: Ordered action
  tokenization,'' \emph{arXiv preprint arXiv:2602.04215}, 2026.

\bibitem{vqbet}
S.~Lee, Y.~Wang, H.~Etukuru, H.~J. Kim, N.~M.~M. Shafiullah, and L.~Pinto,
  ``Behavior generation with latent actions,'' in \emph{Proc. International
  Conference on Machine Learning}, 2024, pp. 26\,991--27\,008.

\bibitem{vqvla}
Y.~Wang, H.~Zhu, M.~Liu, J.~Yang, H.-S. Fang, and T.~He, ``{VQ-VLA}: Improving
  vision-language-action models via scaling vector-quantized action
  tokenizers,'' in \emph{Proc. IEEE/CVF International Conference on Computer
  Vision}, 2025, pp. 11\,089--11\,099.

\bibitem{actioncodec}
Z.~Dong, Y.~Liu, S.~Zhang, B.~Ye, Y.~Yuan, F.~Ni, J.~Gong, X.~Qiu, H.~Zhao,
  Y.~Li, and J.~Hao, ``{ActionCodec}: What makes for good action tokenizers,''
  \emph{arXiv preprint arXiv:2602.15397}, 2026.

\bibitem{xtokenizer}
M.~Kang, L.~Shi, L.~Liang, R.~Gan, D.~Liu, P.~Zhang, S.~Chen, S.~Qin, Y.~Zheng,
  J.~Zheng, H.~Wang, X.~Zhan, and H.~Su, ``{X-Tokenizer}: A multimodal action
  tokenizer for vision-language-action pretraining,'' \emph{arXiv preprint
  arXiv:2606.14752}, 2026.

\bibitem{playlmp}
C.~Lynch, M.~Khansari, T.~Xiao, V.~Kumar, J.~Tompson, S.~Levine, and
  P.~Sermanet, ``Learning latent plans from play,'' in \emph{Proc. Conference
  on Robot Learning}, 2020, pp. 1113--1132.

\bibitem{quest}
A.~Mete, H.~Xue, A.~Wilcox, Y.~Chen, and A.~Garg, ``{QueST}: Self-supervised
  skill abstractions for learning continuous control,'' in \emph{Advances in
  Neural Information Processing Systems}, vol.~37, 2024, pp. 4062--4089.

\bibitem{prise}
R.~Zheng, C.-A. Cheng, H.~Daum\'{e}, III, F.~Huang, and A.~Kolobov, ``{PRISE}:
  {LLM}-style sequence compression for learning temporal action abstractions in
  control,'' in \emph{Proc. International Conference on Machine Learning},
  2024, pp. 61\,267--61\,286.

\bibitem{tcc}
D.~Dwibedi, Y.~Aytar, J.~Tompson, P.~Sermanet, and A.~Zisserman, ``Temporal
  cycle-consistency learning,'' in \emph{Proc. IEEE/CVF Conference on Computer
  Vision and Pattern Recognition}, 2019, pp. 1801--1810.

\bibitem{lav}
S.~Haresh, S.~Kumar, H.~Coskun, S.~N. Syed, A.~Konin, Z.~Zia, and Q.-H. Tran,
  ``Learning by aligning videos in time,'' in \emph{Proc. IEEE/CVF Conference
  on Computer Vision and Pattern Recognition}, 2021, pp. 5548--5558.

\bibitem{softdtw}
M.~Cuturi and M.~Blondel, ``Soft-{DTW}: A differentiable loss function for
  time-series,'' in \emph{Proc. International Conference on Machine Learning},
  2017, pp. 894--903.

\bibitem{fsq}
F.~Mentzer, D.~Minnen, E.~Agustsson, and M.~Tschannen, ``Finite scalar
  quantization: {VQ-VAE} made simple,'' \emph{arXiv preprint arXiv:2309.15505},
  2023.

\bibitem{rt1}
A.~Brohan, N.~Brown, J.~Carbajal, Y.~Chebotar, J.~Dabis, C.~Finn,
  K.~Gopalakrishnan, K.~Hausman, A.~Herzog, J.~Hsu, J.~Ibarz, B.~Ichter,
  A.~Irpan, T.~Jackson, S.~Jesmonth, N.~J. Joshi, R.~Julian, D.~Kalashnikov,
  Y.~Kuang, I.~Leal, K.-H. Lee, S.~Levine, Y.~Lu, U.~Malla, D.~Manjunath,
  I.~Mordatch, O.~Nachum, C.~Parada, J.~Peralta, E.~Perez, K.~Pertsch,
  J.~Quiambao, K.~Rao, M.~Ryoo, G.~Salazar, P.~Sanketi, K.~Sayed, J.~Singh,
  S.~Sontakke, A.~Stone, C.~Tan, H.~Tran, V.~Vanhoucke, S.~Vega, Q.~Vuong,
  F.~Xia, T.~Xiao, P.~Xu, S.~Xu, T.~Yu, and B.~Zitkovich, ``{RT-1}: Robotics
  transformer for real-world control at scale,'' \emph{arXiv preprint
  arXiv:2212.06817}, 2022.

\bibitem{vqvae}
A.~van~den Oord, O.~Vinyals, and K.~Kavukcuoglu, ``Neural discrete
  representation learning,'' in \emph{Advances in Neural Information Processing
  Systems}, vol.~30, 2017.

\bibitem{act2vec}
G.~Tennenholtz and S.~Mannor, ``The natural language of actions,'' in
  \emph{Proc. International Conference on Machine Learning}, 2019, pp.
  6196--6205.

\bibitem{dyne}
W.~F. Whitney, R.~Agarwal, K.~Cho, and A.~Gupta, ``Dynamics-aware embeddings,''
  in \emph{Proc. International Conference on Learning Representations}, 2020.

\bibitem{tcn}
P.~Sermanet, C.~Lynch, Y.~Chebotar, J.~Hsu, E.~Jang, S.~Schaal, and S.~Levine,
  ``Time-contrastive networks: Self-supervised learning from video,''
  \emph{arXiv preprint arXiv:1704.06888}, 2017.

\bibitem{unified}
G.~Chen, Q.~Shao, T.~Cui, Z.~Zhou, W.~Mao, L.~Yang, M.~Wang, Y.~Yang, H.~Chen,
  and Y.~Yue, ``Learning a unified latent action space from videos with
  action-centric cycle consistency,'' in \emph{Proc. IEEE/CVF Conference on
  Computer Vision and Pattern Recognition}, 2026.

\bibitem{hilam}
H.~Kim, L.~Pinto, and S.~J. Kim, ``Hierarchical latent action model,''
  \emph{arXiv preprint arXiv:2603.05815}, 2026.

\bibitem{dp}
C.~Chi, Z.~Xu, S.~Feng, E.~Cousineau, Y.~Du, B.~Burchfiel, R.~Tedrake, and
  S.~Song, ``Diffusion policy: Visuomotor policy learning via action
  diffusion,'' \emph{arXiv preprint arXiv:2303.04137}, 2023.

\bibitem{octo}
{Octo Model Team}, D.~Ghosh, H.~Walke, K.~Pertsch, K.~Black, O.~Mees,
  S.~Dasari, J.~Hejna, T.~Kreiman, C.~Xu, J.~Luo, Y.~L. Tan, L.~Y. Chen,
  P.~Sanketi, Q.~Vuong, T.~Xiao, D.~Sadigh, C.~Finn, and S.~Levine, ``{Octo}:
  An open-source generalist robot policy,'' \emph{arXiv preprint
  arXiv:2405.12213}, 2024.

\bibitem{dtw}
H.~Sakoe and S.~Chiba, ``Dynamic programming algorithm optimization for spoken
  word recognition,'' \emph{IEEE Transactions on Acoustics, Speech, and Signal
  Processing}, vol.~26, no.~1, pp. 43--49, 1978.

\bibitem{ddpm}
J.~Ho, A.~Jain, and P.~Abbeel, ``Denoising diffusion probabilistic models,'' in
  \emph{Advances in Neural Information Processing Systems}, vol.~33, 2020, pp.
  6840--6851.

\bibitem{ste}
Y.~Bengio, N.~L\'{e}onard, and A.~Courville, ``Estimating or propagating
  gradients through stochastic neurons for conditional computation,''
  \emph{arXiv preprint arXiv:1308.3432}, 2013.

\bibitem{film}
E.~Perez, F.~Strub, H.~de~Vries, V.~Dumoulin, and A.~Courville, ``{FiLM}:
  Visual reasoning with a general conditioning layer,'' in \emph{Proc. AAAI
  Conference on Artificial Intelligence}, vol.~32, 2018.

\bibitem{libero}
B.~Liu, Y.~Zhu, C.~Gao, Y.~Feng, Q.~Liu, Y.~Zhu, and P.~Stone, ``{LIBERO}:
  Benchmarking knowledge transfer for lifelong robot learning,'' \emph{arXiv
  preprint arXiv:2306.03310}, 2023.

\bibitem{robocasa}
S.~Nasiriany, A.~Maddukuri, L.~Zhang, A.~Parikh, A.~Lo, A.~Joshi, A.~Mandlekar,
  and Y.~Zhu, ``{RoboCasa}: Large-scale simulation of everyday tasks for
  generalist robots,'' in \emph{Proc. Robotics: Science and Systems}, 2024.

\bibitem{robotwin}
T.~Chen, Z.~Chen, B.~Chen, Z.~Cai, Y.~Liu, Z.~Li, Q.~Liang, X.~Lin, Y.~Ge,
  Z.~Gu, W.~Deng, Y.~Guo, T.~Nian, X.~Xie, Q.~Chen, K.~Su, T.~Xu, G.~Liu,
  M.~Hu, H.~ang Gao, K.~Wang, Z.~Liang, Y.~Qin, X.~Yang, P.~Luo, and Y.~Mu,
  ``{RoboTwin 2.0}: A scalable data generator and benchmark with strong domain
  randomization for robust bimanual robotic manipulation,'' \emph{arXiv
  preprint arXiv:2506.18088}, 2025.

\bibitem{transformer}
A.~Vaswani, N.~Shazeer, N.~Parmar, J.~Uszkoreit, L.~Jones, A.~N. Gomez,
  {\L}.~Kaiser, and I.~Polosukhin, ``Attention is all you need,'' in
  \emph{Advances in Neural Information Processing Systems}, vol.~30, 2017.

\bibitem{flowmatching}
Y.~Lipman, R.~T.~Q. Chen, H.~Ben-Hamu, M.~Nickel, and M.~Le, ``Flow matching
  for generative modeling,'' in \emph{Proc. International Conference on
  Learning Representations}, 2023.

\bibitem{ki}
D.~Driess, J.~T. Springenberg, B.~Ichter, L.~Yu, A.~Li-Bell, K.~Pertsch, A.~Z.
  Ren, H.~Walke, Q.~Vuong, L.~X. Shi, and S.~Levine, ``Knowledge insulating
  vision-language-action models: Train fast, run fast, generalize better,''
  \emph{arXiv preprint arXiv:2505.23705}, 2025.

\end{thebibliography}
\end{document}